# Seeking the Truth Beyond the Data. An Unsupervised Machine Learning Approach

D Saligkaras[b)] and V E Papageorgiou[a)]

*Department of Mathematics, Aristotle University of Thessaloniki, Thessaloniki, Greece*

[a)] Corresponding author: vpapageor@math.auth.gr
[b)] saligkarasd@gmail.com

**Abstract.** Clustering is an unsupervised machine learning methodology where unlabeled elements/objects are grouped together aiming to the construction of well-established clusters that their elements are classified according to their similarity. The goal of this process is to provide a useful aid to the researcher that will help the identification of patterns among the data. Dealing with large databases, such patterns may not be easily detectable without the contribution of a clustering algorithm. This article provides a deep description of the most widely used clustering methodologies accompanied by useful presentations concerning suitable parameter selection and initializations. Simultaneously, this article not only represents a review highlighting the major elements of examined clustering techniques but emphasizes the comparison of these algorithms' clustering efficiency based on 3 datasets, revealing their existing weaknesses and capabilities through accuracy and complexity, during the confrontation of discrete and continuous observations. The produced results help us extract valuable conclusions about the appropriateness of the examined clustering techniques in accordance with the dataset's size.

## INTRODUCTION

One of the human abilities is the object classification into different groups, depending on the characteristics that identify them. With the advent of science, it has become clear that the classification process is necessary to study living things and other objects more systematically. We know that since ancient times, Aristotle had created a system for species categorization in the animal kingdom, which had as its starting point the distinction of animals into vertebrates and invertebrates. This effort in parallel with Theophrastus's attempt on plants, were perhaps the first attempts in the field of taxonomy under the science of Biology. In the science of Geology, it is considered that the material of the meteors is also the primary material of the solar system, that is, material free from alterations and transformations to which the materials of the planets underwent during their formation. Thus, in order to better understand the information coming from the meteors, Geophysics classifies them at an early stage, into four main categories depending on the content of the samples in silicate materials and metals [32]. Other attempts even refer to epidemiology and COVID-19, where a year after its onset, the virus has infected >100 million people [29].

Having mentioned the above, a classification attempt can literally concern everything. There are various algorithmic tools used for classification purposes. There are algorithms that fall into the category of supervised or unsupervised learning. The main difference between these two groups, is that in the first case we know in advance the available classes within each object should be assigned [27]. A subset of the original dataset is used to train the algorithm, enabling the classification outside the training set with the best possible accuracy [30]. In the case of unsupervised learning, no labels are assigned to the examined objects [16]. A methodology that is widely used for these purposes is Cluster Analysis (Clustering), which is a process of categorizing objects, aiming to achieve maximum similarity between the elements that belong into the same cluster.

In recent decades the problem of Cluster Analysis has been thoroughly studied and various algorithms have been proposed, while the way that the clustering procedure is approached differs greatly from algorithm to algorithm. Fraley and Raftery (2002) [8] initially proposed the separation of the various clustering techniques into hierarchical and partitioning. Although, there are other methodologies that are based on the density of observations or the application of statistical models, resulting in the addition of more clustering categories to the already existed ones.

In this article, we explore the efficiency of the most widely used clustering techniques evaluated on datasets containing discrete and/or continuous observations. Moreover, we provide important characteristics about the examined algorithms concerning their operation, suitable parameter selection and appropriate initializations. Through this part of analysis, we are able to extract valuable conclusions not only for the accuracy and the convergence time of these methods, but also for their suitability according to the nature and the structure of the utilized dataset.

The paper is organized as follows; in section 2 we present the related work that exists in the cluster analysis field. In section 3, we outline the major characteristics in parallel with the algorithm's steps for each of the selected grouping techniques. In section 4, we display the efficiency of the examined algorithms applied on 3 simulated datasets containing continuous or/and discrete observations in the 2-dimensional plane. Finally, in sections 5 and 6 we discuss and summarize the main conclusions of the present analysis, focusing on each method's accuracy and convergence time.

## RELATED WORK

Many articles so far have tried to summarize the extensive literature that exists around the field of cluster analysis. Characteristic instances of this studies are the articles of Namratha (2012) [24], Wegmann et al. (2021) [40], Kumar (2016) [17] and Omran et al. (2007) [25]. Furthermore, Gupta et al. (2018) [11] and Gu (2021) [9] made evaluations and comparisons between clustering algorithms but only for two or three cases while in both surveys the algorithmic running time was not measured in order to extract useful conclusions for the algorithms' efficiency. Liao in 2005 [19] published a review paper of clustering techniques for timeseries data. The majority of relevant articles, try to present only the operation of examined algorithms without any attempt of verifying their accuracy and speed of convergence to conclude into some possible comparisons between them. In other cases, a simple reference is made to the worst-case running time of each algorithm, although without attempting to examine the actual convergence times for specific reference datasets.

There are many methods that have been proposed to evaluate the clustering's quality. Such methods are the silhouette method, the similarity table [35], and the Hopkins statistic. But there is a more obvious way to test the capabilities and the accuracy of an algorithm before using the previous methods. The creation of artificial datasets – where there are pre-existing structures with easily established limits – enables the efficient accuracy estimation, highlighting the trustworthiness of the latter comparisons. Thus, this article provides valuable information about the compatibility and capability of each algorithm based on different sets, while providing a representative illustration of the algorithms' application to real-life datasets. Finally, we culminated in the selection of 2-dimensional observations, due to the convenience they offer in the visual examination of the clusters, while we should not trust algorithms that fail to correctly classify 2-dimensional elements, especially when we aim to draw conclusions about higher dimensional data.

## MATERIALS AND METHODS

### Hierarchical Clustering Methods

In hierarchical methods, clusters are formed using either the bottom-up or top-down approach, while the two most widely-used forms of this clustering technique are the agglomerative and the divisive ones. Agglomerative methods follow the bottom-up approach according to which the initially created clusters contain only one element. During each step of the process, the initial clusters are merging, resulting into groups that contain more and more elements. The process is terminated until all objects are placed in a single cluster or other supplementary convergence conditions are satisfied. Divisive methods follow a top-down approach that gradually splits an initial cluster containing all objects into smaller and smaller clusters until each object represents a different group. Hierarchical methods usually lead to the creation of tree diagrams that aid the extraction of useful conclusions concerning the selection of the appropriate number of clusters [23].

The decision of which clusters should be joined or separated is usually defined based on a similarity measure, which combines elements' distance with a linkage criterion that determines the similarity of two groups. The three most common linkage criteria between two groups *A* and *B* are [14]:

1. Complete-linkage; defined as $\max\{d(x,y): x \in A, y \in B\}$ and displaying the maximum distance between the elements of classes $A$ and $B$.
2. Single-linkage; defined as $\min\{d(x,y): x \in A, y \in B\}$ and displaying the minimum distance between the elements of classes $A$ and $B$.
3. Average-linkage; defined as $\frac{1}{|A||B|}\sum_{x \in A}\sum_{y \in B} d(x,y)$ and displaying the average distance between the elements of classes $A$ and $B$.

Except from the above methods, there are other approaches such as the Ward's, the Median and the Weighted Group Average method. The termination criterion for this process – that determines which part of the tree diagram must be kept before all the elements are gathered in one cluster (agglomerative approach) or before they are all divided into classes of one object (divisive approach) – is set by the user. In other words, the process stops when the number of classes are sufficiently large (divisive approach) or sufficiently small (agglomerative approach).

It would be nice to have a relationship that can encapsule a large group of hierarchical algorithms into a family. Lance and Williams 1967 [18], solved this problem by managing to encapsule a wide range of hierarchical methodologies, proposing an equation that with proper adjustments can represent any hierarchical model. At each iterative step, the distance of a selected cluster from all the remaining ones is calculated aiming to aid the continuation of the procedure by finding the next classes that should be merged, until a termination criterion is satisfied.

*Single Link Method*

It is one of the most common hierarchical methods and was first proposed in 1951 by Florek [3] and independently by Sneath [38] in 1957. The distance of two classes is the smallest distance provided by their objects. Hence:

$$D(C, C') = \min_{x \in C, y \in C'} d(x, y), \tag{1}$$

where $d$ is often considered as the Euclidean Distance. Let $C_i$, $C_j$, $C_k$ be 3 classes and $C_i$, $C_j$ should be merged. Then, the distance of the new class $C_i \cup C_j$ from $C_k$ is calculated as:

$$D(C_i \cup C_j, C_k) = \tfrac{1}{2}D(C_i, C_k) + \tfrac{1}{2}D(C_j, C_k) - \tfrac{1}{2}|D(C_i, C_k) - D(C_j, C_k)| = \min\left(D(C_k, C_i), D(C_k, C_j)\right). \tag{2}$$

*Complete Link Method*

In contrast to the previous method, the distance between $C_i$ and $C_j$ is considered to be the largest distance between the elements of these two classes. Now, the distance of the merged class $C_i \cup C_j$ from $C_k$ is calculated as:

$$D(C_i \cup C_j, C_k) = \tfrac{1}{2}D(C_i, C_k) + \tfrac{1}{2}D(C_j, C_k) + \tfrac{1}{2}|D(C_i, C_k) - D(C_j, C_k)| = \max\left(D(C_k, C_i), D(C_k, C_j)\right), \tag{3}$$

where:

$$D(C, C') = \max_{x \in C, y \in C'} d(x, y). \tag{4}$$

*Group Average Method*

This method is also called UPGMA (Unweighted Pair Group Method using arithmetic Average). It is defined as the average distance of all pairs of the two examined classes. The distance of the new merged class from class $C_k$, is calculated as:

$$D(C_i \cup C_j, C_k) = \frac{|C_i|}{|C_i|+|C_j|}D(C_i, C_k) + \frac{|C_j|}{|C_i|+|C_j|}D(C_j, C_k), \tag{5}$$

where for $C$ and $C'$ we have:

$$D(C, C') = \frac{1}{|C||C'|}\sum_{x\in C, y\in C'} d(x,y). \tag{6}$$

*Ward's Method*

In 1963, Ward proposed a method that aims to minimize a function-criterion whose value changes during the algorithm's accumulation process [39]. This function is the total sum of the squares of the distances from the center $l(C)$. For a dataset $C$ this function is denoted as $ESS(C)$ where:

$$ESS(C) = \sum_{x\in C}(x - l(C))(x - l(C))^T. \tag{7}$$

In addition, we assume that at some point during the accumulation process there are $k$ in number clusters. The total ESS is defined as:

$$ESST = \sum_{i=1}^{k} ESS(C_i). \tag{8}$$

At each step, the union of all blocks is examined and the merging process that produces the lowest $ESST$ is selected. The distance between the objects of a merged cluster $C_i \cup C_j$ from a third one $C_k$, can be calculated using the Lance-Williams equation (Wishart 1969).

The simple agglomerative algorithm has complexity $O(n^3)$. For the single-linkage and complete-linkage methods there are alternatives with complexity $O(n^2)$. As for the Ward's method, a first improvement produced a complexity of $O(n^2 \log(n))$ and a latter one a complexity of $O(n^2)$. Finally, for the UPGMA method, an algorithm of complexity $O(k3^k n^2)$ has been presented by Day et al. [4], while Murtagh gave another improvement of complexity $O(n^2)$ [22].

## K-Means

The original algorithm was first proposed in 1957 by Stuart Lloyd [20], which is an iterative procedure often called Lloyd's algorithm or naive K-means, since other alternatives have been proposed along the way.

Given an initial set of $k$ centers $m_1^1, m_2^1, \ldots, m_k^1$ the algorithm is based on the following steps.

Step 1: Each observation must be included in the cluster of the nearest medoid. The distance used to define proximity is usually the squared Euclidean distance:

$$S_i^t = \left\{x_p: |x_p - m_i^t|^2 \leq |x_p - m_j^t|^2 \; \forall j, 1 \leq j \leq k\right\}, \tag{9}$$

where each observation $x_p$ is inserted only in the cluster $S^t$.

Step 2: The medoids of the new clusters are calculated as:

$$m_i^{t+1} = \frac{1}{|S_i^t|}\sum x_j, \tag{10}$$

where $|S^t|$ denotes the number of elements that have joined the $S_i$ class during the *t*-th iteration. During the operation of the algorithm, we aim to minimize the value of the within clusters sum of squares (WCSS), presented by:

$$wcss = \sum_{i=1}^{k}\sum_{x_j \in S_i}|x_j - m_i|^2. \tag{11}$$

The algorithm converges when each cluster's elements are consolidated. However, there is no guarantee that the best solution will be found. In addition, it is necessary that the number of final clusters should be known a priori. As a result, this procedure leads to diversified results based on the preselected number of clusters [2].

Increasing continuously the number of clusters inevitably leads to an error reduction. For this reason, a balance must be accomplished between over-compressing the dataset into a small number of clusters while minimizing the corresponding sum of squares. This balance is succeeded by observing the diagram of WCSS values for each selection of clusters' number $k$. The ideal selection for $k$, is the value that corresponds to the emergence of a plateau in the $wcss$ graph [28].

Another parameter that needs to be predetermined, is the selection of the centers of the clusters, where the 2 most popular techniques are Forgy and Random Partition [12]. In the case of the Forgy Method, $k$ elements are randomly selected from the initial dataset to constitute the initial centers. In the second case, each dataset's element is randomly assigned to one of the $k$ clusters, while all elements belong to a cluster with the same probability. Note that the initial medoids have not yet been selected. These centers are selected in the second step of the algorithm. The medoid initialization method that leads to faster convergence, depends highly on the nature of the data.

The complexity of the simple K-means is $O(n \times k \times I \times d)$, where $n$ is the number of the dataset's elements, $k$ is the number of clusters, $d$ is the number of attributes that describe each object, and $I$ is the number of iterations that are needed until convergence.

## K-Means ++

Another approach aiming to improve the selection of the $k$ initial medoids, is offered by the K-means ++ algorithm [2]. A similar attempt with this of Arthur and Vassilvitskii was proposed by Ostrovsky et al. in 2006 [26]. This algorithm tries to enhance the clustering performance of K-means that is based on a random selection of initial centers.

For instance, consider four points on the 2-dimensional plane forming a rectangle whose length is considerably greater than its height. Suppose $k = 2$ is chosen and the two initial centers are located, one in the middle of the straight line connecting the two upper vertices and the other in the middle of the line connecting the two bottom ones. The algorithm then converges directly without calculating new centers. Hence, the two upper and the two lower vertices represent the produced clusters. However, this selection does not minimize the $wcss$.

The K-means ++ algorithm works as follows. An object is randomly selected from the dataset to be the first medoid. Each item is selected with the same probability. The remaining objects are selected as medoids with a probability proportional to their distance from the nearest already chosen medoid [2]. This process continues until $k$ centers are selected. In summary, the process consists of the following four steps:

- Step 1: Select a medoid $c_1$, uniformly from dataset X.
- Step 2: Choose a medoid $c_i$ from X with probability $\frac{D(x)^2}{\sum_{x \in X} D(x)^2}$.
- Step 3: Repeat the previous step until $k$ medoids are selected.
- Step 4: Continue as dictated by the K-means algorithm.

Applications have shown that K-means ++ manages to surpass the clustering accuracy of K-means, highlighting the importance of efficient medoid initialization. The selection of the initial center points requires $k$ additional scans of the dataset. However, this process does not significantly increase the running time of K-means ++ compared to the conventional K-means. Namely, in many applications the K-means ++ requires shorter running time intervals for its operation, as a suitable medoid initialization leads to quite faster convergence times.

## PAM

The PAM (Partition Around Medoids) algorithm aims to find objects called medoids located at the center of the clusters. These medoids are placed in a set $S$ called the set of selected objects. If $O$ is the set of all objects, then we define the set of unselected objects $U = O - S$. The aim of the algorithm is to minimize the objects' ambiguity with their nearest selected object, utilizing the Euclidean or in certain occasions the Manhattan distance. The algorithm has two main phases [15].

Phase 1: (BUILD). A collection of $k$ objects is selected to create the initial set $S$.

Phase 2: (SWAP). An effort is made to improve the quality of the initial clustering by exchanging the selected with the non-selected items, where necessary. We define the quantities $D_p$ and $E_p$ that denote the dissimilarity between an object $p$ and it's first and second closest neighbor in the set $S$, correspondingly. The execution of PAM algorithm is based on the following steps.

Step 1: The object with the minimum total distance from all the other objects is placed inside set $S$, representing the most central point of the dataset $O$.

Step 2: An object $i \in U$, is randomly selected as a candidate for membership in $S$.

Step 3: For an object $j \in U - \{i\}$ calculate $D_j$ and if $d(i,j) < D_j$, object $j$ contributes to the decision of selecting object $i$.

Step 4: Consider $C_{ij} = max(D_j - d(i,j), 0)$ and calculate the quantity $g_i = \sum_{j \in U} C_{ij}$. Finally, select object $i$ which maximizes the quantity $g_i$.

Step 5: Sets $S$ and $U$ are updated after each iteration.

These steps are applied until $k$ items are selected for the set $S$. Going to the 2nd phase, we aim to improve the grouping quality. First, we consider all pairs $(i,h) \in S \times U$ and calculate the influence $T_{ih}$ of a possible replacement of an element $i$ with an element $h$ in the set $S$. To calculate $T_{ih}$ one must first calculate the contribution $K_{ijh}$ of each object $j \in U - \{h\}$ during the replacement of $i$ with $h$. The 2hd phase's steps are

Step 1: Calculate $K_{ijh}$ considering the following two cases:

- if $d(j,i) > D_j$ then $K_{ijh} = min(d_{jh} - D_j, 0)$;
- if $d(j,i) = D_j$ then $K_{ijh} = min(d_{jh}, E_j) - D_j$.

Step 2: Calculate $T_{ih} = \sum K_{jih}$, $j \in U$.
Step 3: Select a pair $(i,h) \in S \times U$ that minimizes $T_{ih}$.
Step 4: If $T_{ih} < 0$, execute the replacement of the 2 elements and return to the Step 1.
Step 5: The algorithm terminates when all $T_{ih}$ become positive.

Applications have shown that PAM is efficient for smaller datasets, which is a bit subjective since the computing power of even a personal computer is satisfactory for simulations containing a few thousand observations. However, when we deal with larger datasets, the algorithm's complexity increases significantly, as during the 3rd step, there is a total of $k(n-k)$ pairs and for each pair, the calculation of $T_{ih}$ requires the examination of $n-k$ objects. Therefore, for one and only iteration we have complexity $O(k(n-k)^2)$ [15]. This fundamental disadvantage was the motivation for the development of CLARA algorithm.

## CLARA

CLARA (Clustering LARge Applications) was designed to manage large datasets, in contrast with PAM [15]. To accomplish this, instead of finding representative objects-elements for the whole dataset, this algorithm takes a sample from the set and applies the PAM algorithm to the selected sample finding its medoids. The philosophy of this process emphasizes that if the sample is taken at random (simple random sampling), the medoids of the sample approach satisfactorily the medoids of the initial dataset. To enhance the algorithm's accuracy, the sampling process is performed several times, while as output we select the grouping that minimizes the mean difference of dataset's elements; the mean difference is calculated by the sum of distances of all points from their nearest medoid. Applications have shown that the selection of 5 samples of $40 + 2k$ items provides reliable results. Briefly, the CLARA algorithm can be summarized in the following steps:

Step 1: For $i = 1, \ldots, L$, choose a sample of $40 + 2k$ objects by simple random sampling and then perform the PAM algorithm to find the $k$ medoids for the sample.

Step 2: For each item of the set $U$, determine the most similar medoid.

Step 3: Calculate the mean difference of the dataset. If it is smaller than the already existed ones, set the new difference as the new minimum and consider the centers that have been found in step 2 as the representative set.

Step 4: Return to Step 1 for the next iteration.

The algorithm returns the clusters resulting from the sample with the lowest mean heterogeneity. Note that the total of non-selected items refers to all items that were not selected during the sampling phase.

As mentioned earlier the complexity of PAM for one iteration is $O(k(n-k)^2)$ where $k$ is the number of groups and $n$ is the total number of elements. For CLARA a random sample $n' = 40 + 2k$ is selected and then PAM is applied for $n = n'$. For each non-selected object, it is necessary to find the nearest medoid. Therefore, additional $k(n-k)$ tests are required. Hence, CLARA's complexity approaches $O(k(40+k)^2 + k(n-k)) \cong O(k^3 + nk)$ which is quite satisfactory for a relatively moderate number of clusters [33].

# DBSCAN

DBSCAN (Density based clustering of applications with noise) was proposed by Ester et al. in 1996 [6] and its philosophy is based on the observations' density. More specifically, the algorithm places together observations that have many neighboring points, while pointing out elements as irregularities that their nearest neighbors are distant. Consider a set of points of some dimension. To define the neighborhood around a reference point it is necessary to define a positive number. Any other point located within the supersphere centered on the reference point with radius $\varepsilon$ is considered an adjacent point. Also, the quantity $minPts$ must be specified, which indicates the minimum number of points for which an area is considered dense.

We divide the dataset's items into categories. A point $p$ is called a core point if in a radius $\varepsilon$ around it, there are at least $minPts$ points including $p$. A point $q$ is called directly reachable from a core point $p$ if the distance of $q$ from p, is less than $\varepsilon$. A point $q$ is called a reachable point from another point $p$ if there is a path $p_1, \ldots p_n$ with $p_1 = p$ and $p_n = q$ while every $p_{i+1}$ is directly reachable from $p$. Also, $p$ is tightly connected to $q$, if there is an element $o$ such that both $p$ and $q$ are close approximations of $o$. Outliers/noise/irregularities are the dataset's elements that do not fall into any of the above categories [6].

Then, if $p$ is a core point, it forms a cluster together with the points that represent its approximations; whether these points are core points or not. Namely, the algorithm selects a starting point $p$ arbitrarily and scans its neighborhood. If the neighborhood is found to be sufficiently dense then the formation of a cluster $C$ initiates, including all the points belonging in the neighborhood of $p$. Then, the neighborhood of all points belonging to this cluster is checked. If the neighborhood for any of these points – let such a point called $q$ – contains $\geq minPts$ points, then all the neighbors of $q$ that have not been already included in cluster $C$ are added. The same process continues until there are no new additions.

If a point cannot be defined as a core point, then it is initially marked as an outlier. At this stage the point is excluded as a possible core and cannot initiate a cluster's creation. Although, this does not mean that it is prohibited of finding itself as part of another cluster since this point can then be considered as directly reachable from another starting point of the algorithm. After the DBSCAN completes the iterative process, it characterizes the remaining points as noise.

Let $d$ be the distance of a point $p$ from it's $k$-th closest neighbor. Then, the $d$ neighborhood $Nd(\epsilon)$ of $p$ contains exactly $k + 1$ points, except there are more points with the same distance $d$ from $p$. For a given value of $k$, we define the $k - dist$ function which corresponds each observation point to the value of its distance from the nearest neighbor $k$. These points are ordered in descending order of $k - dist$ value, resulting in the creation of a declining curve.

Consider the value $\varepsilon$ to be equal to $k - dist$ $(p)$ for some arbitrary point $p$ while $minPts = k$. We aim into finding the curve's part that displays a plateau. The points corresponding to the left part of the plateau are deemed as outliers, while the right ones should be added in the clusters. In practice, various applications have shown that the $k - dist$ graphs for $k > 4$ do not differ significantly from the graph where $k = 4$ [6]. Therefore, in the results section we explore the respective graphs for $k = 3, 4, 5$ [36]. For more than 2 dimensions, the empirical rule $minPts = 2d$ can be followed, where $d$ is dataset's dimension.

DBSCAN has some great advantages; it does not require the cluster number to be given as input, while it also detects points that are very different from the rest of the dataset and characterizes them as noise. Furthermore, it can accurately detect clusters with irregular shapes. Cases where the grouping performance is not satisfactory, are clusters that are characterized by varying densities or the density of data from region to region shows significant changes. The complexity of the algorithm in the worst case is $O(n^2)$. The following graphs display an example where DBSCAN detects clusters of irregular shapes while other algorithms such as K-means fail.

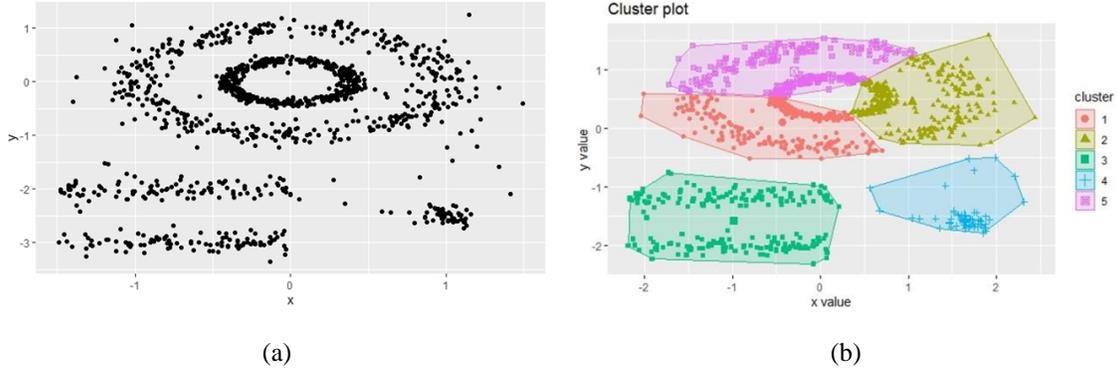

(a) (b)

**FIGURE 1.** Clustering example through the utilization of K-means for $k = 5$.

We notice that the dataset contains five clusters. The two are consisted of objects that form something like concentric circles at the top of the graph. In the lower left part two more clusters are created and finally in the lower right part we meet the fifth cluster. Moreover, we observe objects that are relatively away from any cluster and may represent irregularities.

We apply the K-means algorithm with an input of $k = 5$. In figure 1, it appears that K-means fails to separate the two groups of objects at the top of the graph and instead creates a total of three clusters. The two groups of elements in the lower left part of the figure are merged into one cluster. The fifth group seems to have been successfully identified, including many irregular points. Before applying DBSCAN we procced to the evaluation of the parameter $\varepsilon$ (Figure 2), where we choose $\varepsilon = 0.15$ while $minPts = 5$. The results for $minPts = 4$ do not differ.

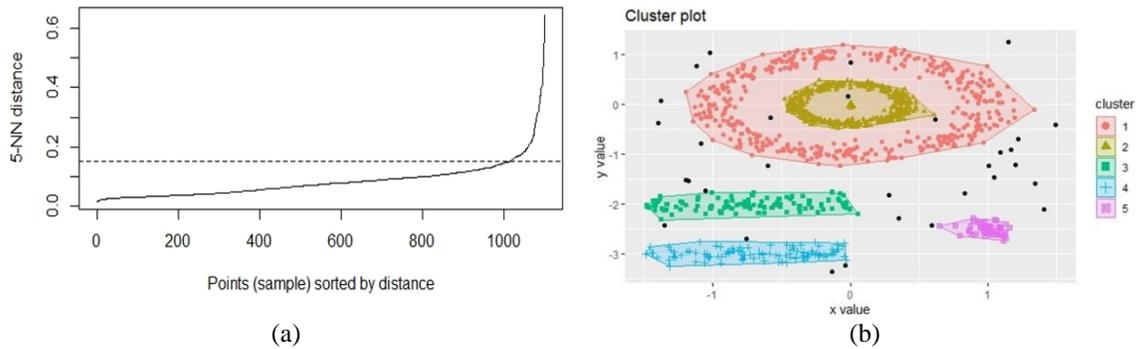

(a) (b)

**FIGURE 2.** Clustering example through the utilization of DBSCAN.

Figure 2 shows that DBSCAN meet the structure expectations, where the separation of the five groups is adequately achieved while a small number of points are labeled as noise (black dots).

## OPTICS

OPTICS (Ordering Points To Identify the Clustering Structure) examines data density [1], while its philosophy is similar to that of DBSCAN but aims to overcome the latter's obstacles. This problem is the creation of groups, considering that the data's density differs significantly from region to region in the observation space. Furthermore, while DBSCAN in most cases is deemed effective, it entrusts the user with the choice of major initial parameters where their manual estimation can be highly unreliable in certain occasions.

To overcome this difficulty, OPTICS operates equivalently with DBSCAN for a set of parameters instead of a specified parameter value, removing the requirement of giving a specific a priori value to $\varepsilon$. In this case, $\varepsilon$ is usually defined as the maximum possible distance from the $minPt$ nearest neighbor. Due to the high value of $\varepsilon$, parameter $minPt$ has less impact in the grouping process and is often set to 5.

The selection of the number of clusters is now determined by a two-dimensional graph, where the observations are ordered on the axis of the independent variable. On the other hand – as dependent variable – the reachability distance is utilized. Now, the core distance describes the distance of each point from its $minPt$ closest point.

$$core\_dist_{minPts}(p) = \begin{cases} Undefined, & if\ |N(p,\varepsilon)| < minPts \\ minPts_{Nearest\_dist}, & else \end{cases}, \quad (12)$$

where $|N(p,\varepsilon)|$ is the number of points in the neighbourhood of $p$. The reachability distance ($r\_dist$) of a point $q$ from a point $p$ is the maximum among its core and their Euclidean distance. Hence, the $r\_dist$ function is defined as:

$$r\_dist_{\varepsilon,minPts}(p) = \begin{cases} Undefined, & if\ |N(p,\varepsilon)| < minPts \\ \max(core\_dist_{\varepsilon-minPts}(p), dist(q,p)), & else \end{cases}. \quad (13)$$

In the final two-dimensional graph, we observe valleys indicating the existence of clusters and peaks – high reachability distance values – indicating the boundaries between clusters. In fact, the larger the valleys, the denser the respective cluster and the lower the radius requirements are.

## Gaussian Mixture Models

Another way to group a dataset's elements is through the utilization of their distribution. In many cases there are elements that do not follow a single distribution. Thus, objects generally originate from different distributions that are often considered to be normal. A sum of these distributions can reliably describe the data distribution when the required parameters are properly estimated.

Quite important elements of this algorithm are the weights that describe the portion that each individual component participates in the mixture-model. Since we assume Gaussian distributions, the estimated parameters are vectors representing the mean values, and coefficient tables that calculate the correlations between the characteristics of the random variables. A common way to estimate these parameters is to use the Expectation-Maximization algorithm. Therefore, we assume $k$ Gaussian distributions with probability density functions $f_1(x_1), \ldots, f_k(x_k)$ where:

$$f_i(x_i) = \left(\frac{1}{2\pi}\right)^{\frac{p}{2}} |\det(\Sigma_i)|^{-\frac{1}{2}} \exp\left(-\frac{1}{2}(x_i - m_i)^T \Sigma_i^{-1}(x_i - m_i)\right). \quad (14)$$

Forming a mixed distribution with a probability density function:

$$f(x; A) = \sum_{l=1}^{k} \pi_l f_l(x; \theta_l), \quad (15)$$

where $A = \{\pi_1, \ldots, \pi_k, \theta_1, \ldots, \theta_k\}$ and $\theta_l = (m_l, \Sigma_l)$.

Considering $N$ independent and identically distributed (i.i.d) random variables that satisfy equation (15), we define the joint probability density function as:

$$L(A) = \prod_{i=1}^{N} \sum_{l=1}^{k} \pi_l f_l(x_i; \theta_l), \quad (16)$$

where $\log(L(A))$ can be written as:

$$\log(L(A)) = \sum_{i=1}^{N} \log\left(\sum_{l=1}^{k} \pi_l f_l(x_i; \theta_l)\right). \quad (17)$$

Furthermore, let $Z$ be an index random variable such that:

$$Z_{kl} = \begin{cases} 1, & if\ x_i\ belongs\ to\ k\ distribution \\ 0, & else \end{cases}. \quad (18)$$

As a result:

$$P(Z_{kl} = 1) = \tau_k^t(x_i; A^t) = \frac{\pi_k^t f_k(x_i; \theta_k^t)}{f(x_i; A^t)} \quad (19)$$

and for the Gaussian mixture model:

$$\tau_l^t(x_i; A^t) = \frac{\pi_l^t N(x_i | m_l^t, \Sigma_l^t)}{\sum_{j=1}^k \pi_j^t N(x_i | m_j^t, \Sigma_j^t)}, \quad (20)$$

where the index $t$ corresponds to the number of iterations. Combining (17) and (20) it appears that [5]:

$$E[\log(L(A))] = \sum_{i=1}^N \sum_{l=1}^k \tau_l^t(x_i; A^t)(\log(\pi_l) + \log(f_l(x_i; \theta_l))). \quad (21)$$

This process concerns the first step of the algorithm (expectation). During the next step we maximize the function (21). The weights $\pi_l^{t+1}$ are estimated independently of the parameters $\theta^{t+1}$. Hence:

$$\pi_l^{t+1} = \frac{1}{N} \sum_{i=1}^N \tau_l^t(x_i; A^t). \quad (22)$$

The estimation of the parameters $\theta^{t+1}$, is based on the system of equations:

$$\sum_{i=1}^N \sum_{l=1}^k \tau_l^t(x_i; A^t) \frac{\partial(\log(f_l(x_i; \theta_l)))}{\partial \theta_l} = 0. \quad (23)$$

It turns out that:

$$m_l^{t+1} = \frac{\sum_{i=1}^N \tau_l^{t+1} x_i}{\sum_{i=1}^N \tau_l^{t+1}(x_i; A^t)} \quad (24)$$

and

$$\Sigma_l^{t+1} = \frac{\sum_{i=1}^N \tau_l^{t+1}(x_i - m_l^{t+1})(x_i - m_l^{t+1})^T}{\sum_{i=1}^N \tau_l^{t+1}(x_i; A^t)}. \quad (25)$$

Commonly we set some arbitrary values to the weights of the 1st step as an initialization prerequisite. The procedure continues until a sufficient convergence occurs between two consecutive iterations. As with most algorithms, the fundamental question that arises is how many clusters should be created. For the GMM, the confrontation of this issue is based on the selection of the adequate number of $k$ Gaussian distributions included in the model. For this purpose, we use the Bayesian Information Criterion (BIC) [5]. As the likelihood increases with the usage of more and more Gaussian Distributions a penalty degree corresponding to the number of estimated parameters decreases the criterion's value. The criterion takes the form:

$$BIC_{M,k} = 2I_{M,k}(x|A) - q\log(n), \quad (26)$$

where $M$ is an index corresponding to a particular model, $q$ is the number of parameters that needs to be estimated and $n$ represents the number of observations. The $k$ corresponding the `most suitable selection, is the one that maximizes the criterion.

## RESULTS

In this section, we explore the required convergence times in parallel with the efficacy of the final clustering results based on 3 reference datasets. We create datasets that their observations are determined by pairs of random variables. Namely, the utilization of artificial datasets is based on the capability they provide to precisely evaluate the quality of the clustering techniques. In this part we explore the performance of K-means, K-means++, PAM, CLARA, DBSCAN, OPTICS, UPGMA and Gaussian Mixed Models.

# Clustering Performance Examined on Mixed Dataset

The first dataset consists of 1000 2-dimensional observations, where the values of the first and second random variables (r.v.) come from a normal and a Poisson distribution respectively, producing four classes with clear boundaries that are not far away from each other. The latter helps us to test the algorithms' capabilities without resulting into a significantly obvious grouping procedure. Thus, using the generators of R software, the data that constitute dataset 1 is displayed in Figure 3.

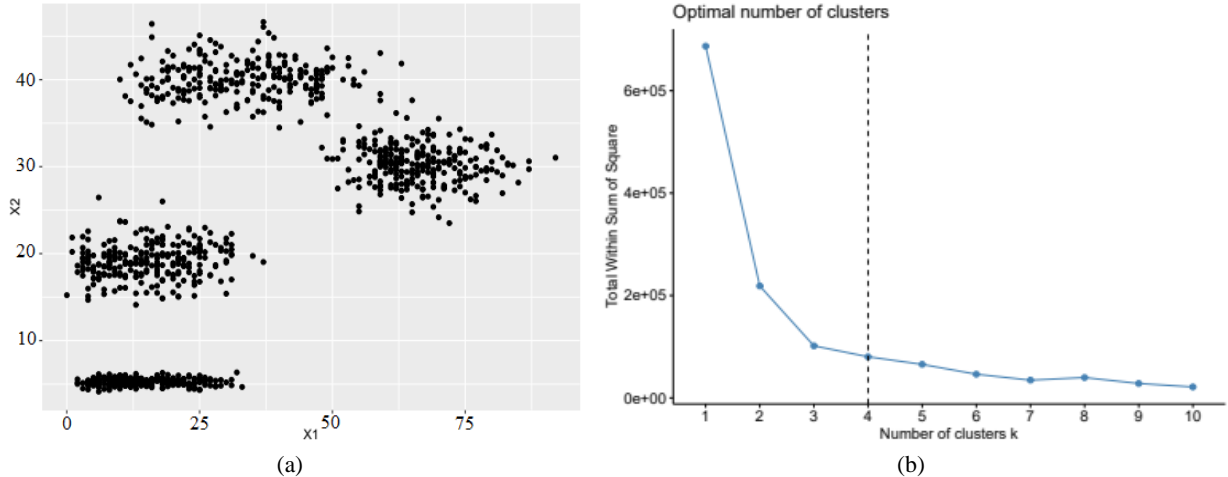

(a)                                                                                           (b)

**FIGURE 3.** Plots for dataset 1 and the respective WCSS diagram, aiding the selection of the appropriate number of clusters.

Using the elbow method, the corresponding graph present a minor decrease after $k = 4$. For a second opinion we may examine the respective reachability plot which is the graph that results from the application of the OPTICS algorithm. The number of valleys that arises, is an indicator for the appropriate number of clusters. For a specific value of the reachability distance, we estimate the suitable values for the parameters $minPts$ and $\varepsilon$, required for the utilization of DBSCAN. We emphasize that OPTICS can be used as a stand-alone tool, to determine the number of clusters and their density. In figure 4, we see the corresponding graph for Dataset 1, observing four high peaks that create four valleys, providing also an indicator for the creation of four clusters. Then, we apply the naïve K-means (Figure 5). It is obvious that due to incorrect selection of the initial centers, the algorithm is unable to separate correctly 2 out of the 4 classes.

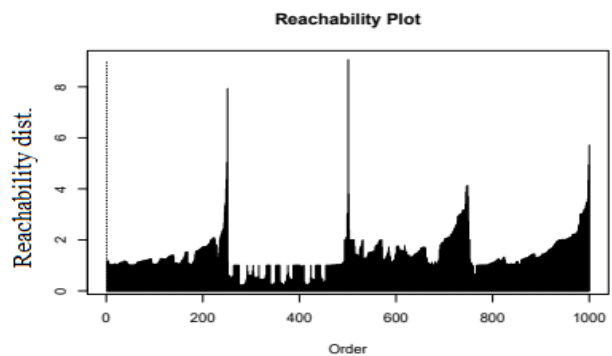

**FIGURE 4.** Reachability plot for Dataset 1.

For input $k = 4$, figure 5 displays the result of K-means++. We notice a more effective object classification compared to the conventional K-means, a phenomenon that derives from the theoretical superiority of K-means++ due to the difference in the selection of initial centers. We continue with the implementation of PAM algorithm considering the above observations for the number of clusters (Figure 6). Compared to K-means the results are significantly better where only a small number of observations has been misclassified.

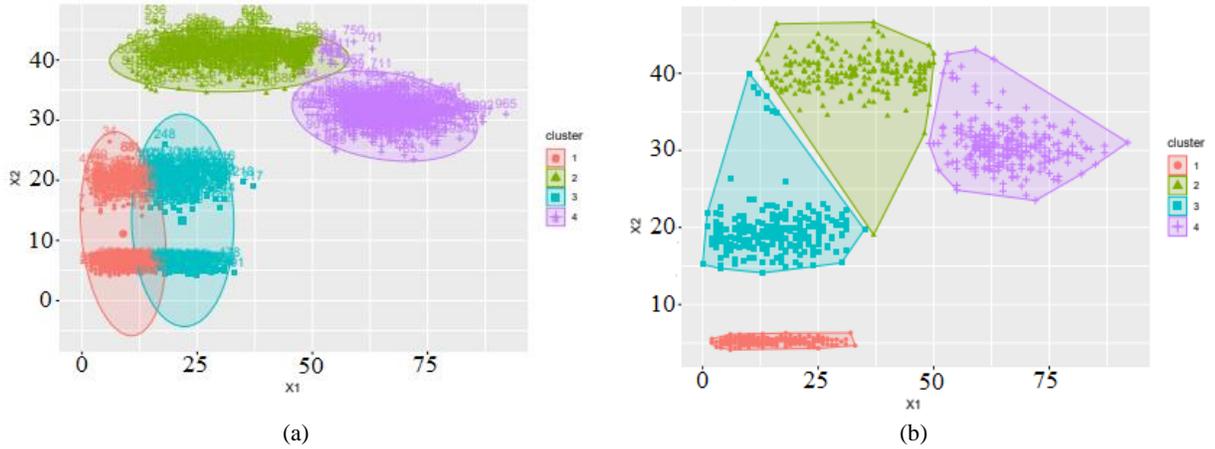

**FIGURE 5.** Clustering results of K-means and K-means++ on dataset 1.

As we previously mentioned, CLARA selects a number of $40 + 2k$ samples and then selects the optimal sample. For the purposes of this application, 48 samples of 50 observations have been used. The large number of chosen samples, results into a satisfactory classification that resembles that of PAM, which is expected, since PAM is the algorithm that creates the clusters in the case of CLARA as well. On the other hand, utilizing the same data in a different implementation of CLARA (with the same number of samples), the outcome is likely to be different.

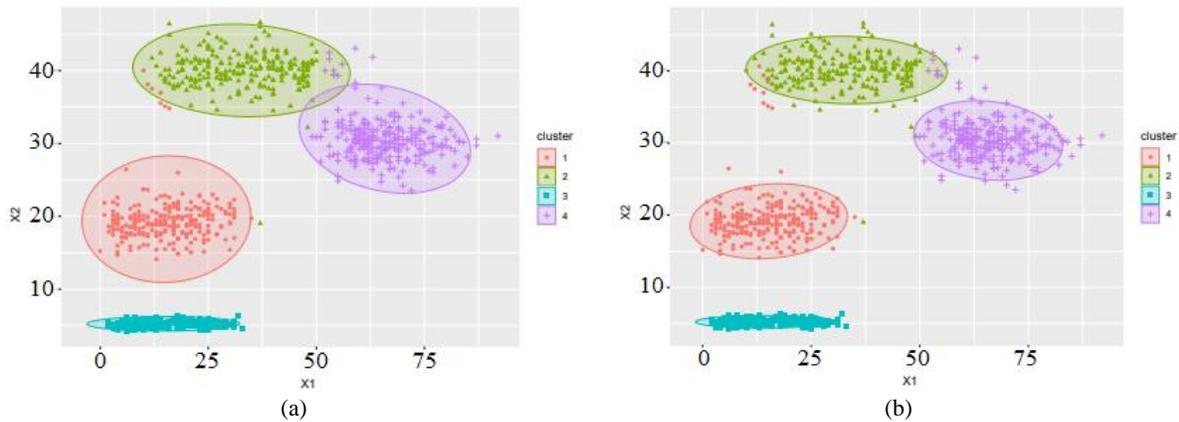

**FIGURE 6.** Clustering results of PAM and CLARA algorithms for dataset 1.

We proceed with the implementation of DBSCAN, where the parameters $minPts$ and $\varepsilon$ have to be evaluated. Figure 7 displays the objects' distances from their 3rd, 4th and 5th closest neighbor respectively, since the improvements observed for increased number of neighbors can be considered insignificant. Thus, for the first two cases of figure 7 the estimation of $\varepsilon$ is shown to be around 2, while according to the third graph we receive an estimation of $\varepsilon = 2.2$.

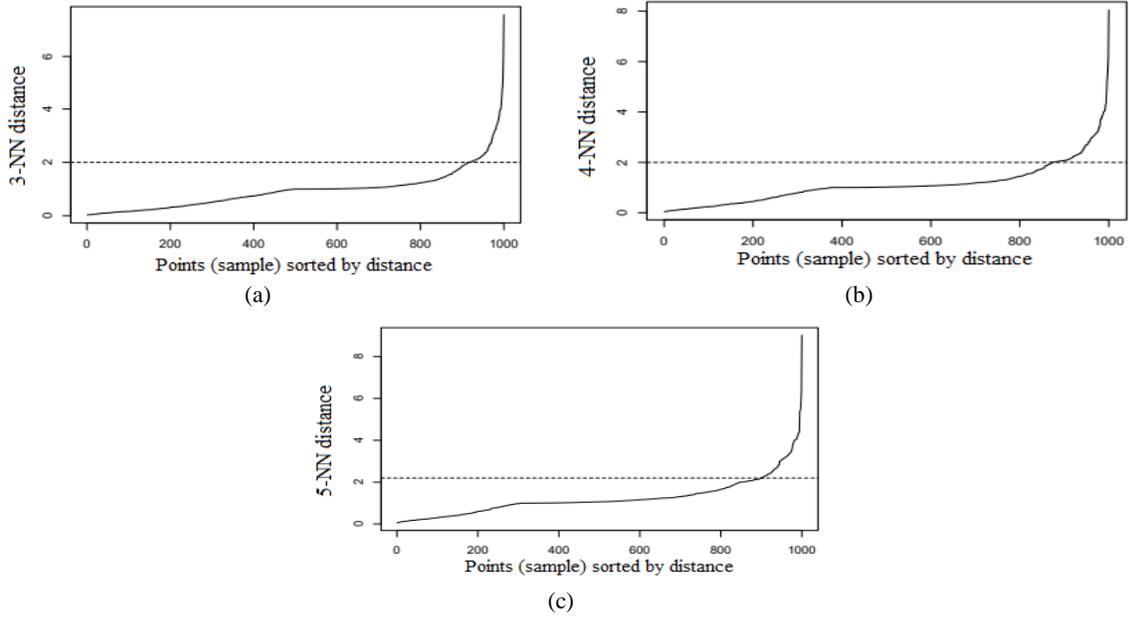

**FIGURE 7.** Graph displaying the distance of points from their 3rd, 4th and 5th nearest neighbor utilized for the estimation of parameter $\varepsilon$.

In figure 8 the respective results for $minPts = 3$ and $\varepsilon = 2$ are displayed. The main body of the four classes is successfully identified. In addition, several elements that are marked as black dots have been considered as irregularities or noise points that are not included in any cluster, while we also observe the creation of extra unnecessary clusters with negligible number of items. We also present the results for $minPts = 4$ and $\varepsilon = 2$, where the observations seem to have been correctly distributed into classes with no significant differences comparatively to the previous instance. Finally, for $\varepsilon = 2.2$ the number of noise points is decreasing, and the percentage of correctly classified elements has been improved.

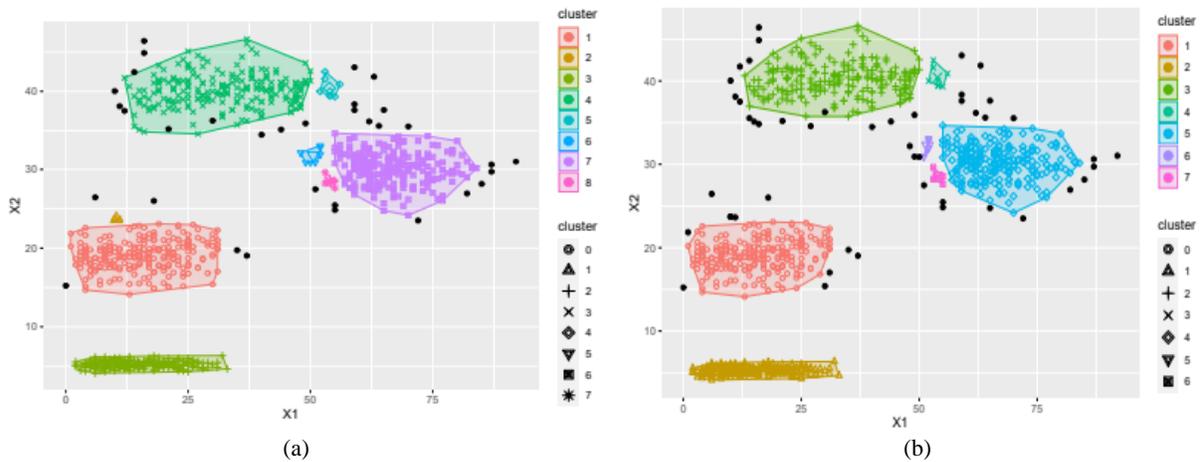

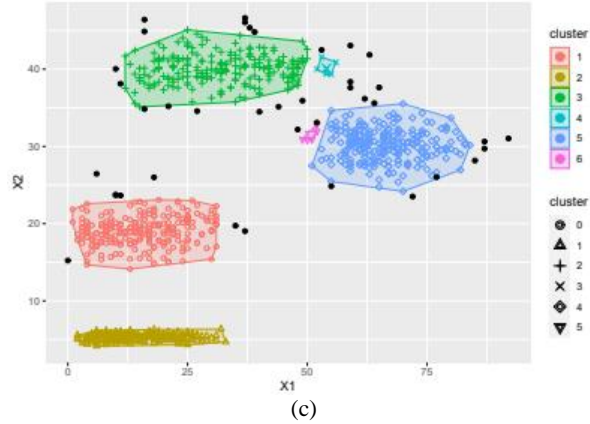

(c)

**FIGURE 8.** DBSCAN clustering for $(minPts, \varepsilon) = (3, 2)$, $(minPts, \varepsilon) = (4, 2)$ and $(minPts, \varepsilon) = (3, 2.2)$ respectively.

The last comment motivates the thought that a larger choice of $\varepsilon$ will further reduce the number of irregularities and eliminate the existence of unwanted clusters of small population. In the reachability plot produced by OPTICS, the horizontal line placed around the lowest of the four highest peaks helps us provide a result with minimal irregular points removing additional classes and enhancing the classification performance. Figure 9 displays the reachability distance plot proposing the utilization of 4 clusters. The elements of each cluster that display a reachability distance value > 4 are the points that will be considered as noise. The very small number of irregular points is confirmed by the OPTICS diagram, while the correspondence of clusters and classes is adequate.

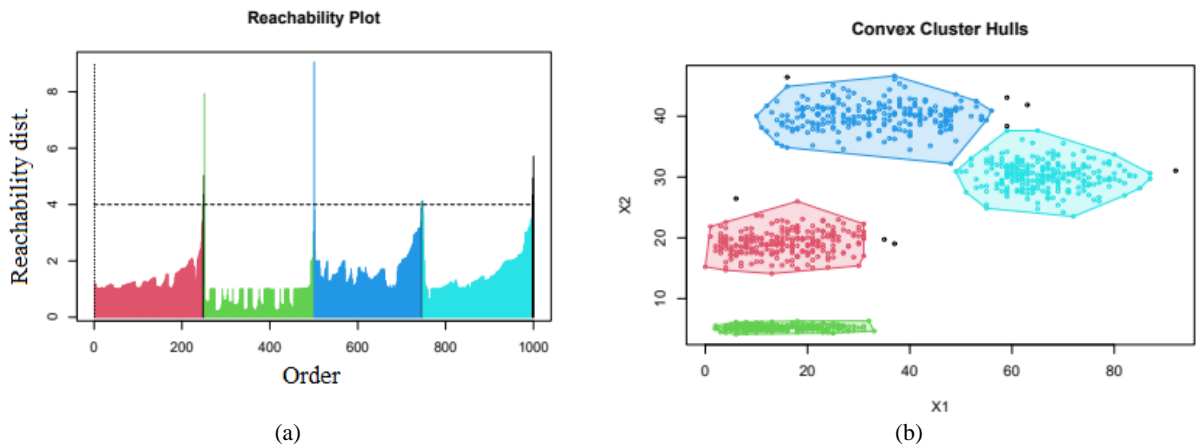

(a)                                                                 (b)

**FIGURE 9.** Reachability distance plot and the clustering results based on OPTICS.

Figure 10 shows the application of BIC (26) on dataset 1. The legend concerns the form of the distributions, such as the area that will be occupied by the ellipsoid and its orientation; information resulting from the estimation of the model's covariance matrix. There is a clear indication that the Gaussian Mixed Model should consist of 4 factors (BIC's plateau) for which the extent and shape of the ellipsoid will differ. The ellipsoid's orientation is based on the X-Y axes. According to the application of the Expectation-Maximization algorithm for 4 factors, figure 10 displays the results for the VVI case, which stands for the diagonal (covariance matrix) and varying volume-shape of each ellipse. More specifically the parameterization is represented by a 4 x 4 diagonal covariance matrix with equal variances and covariances fixed to 0 while the means of each Gaussian vary [37].

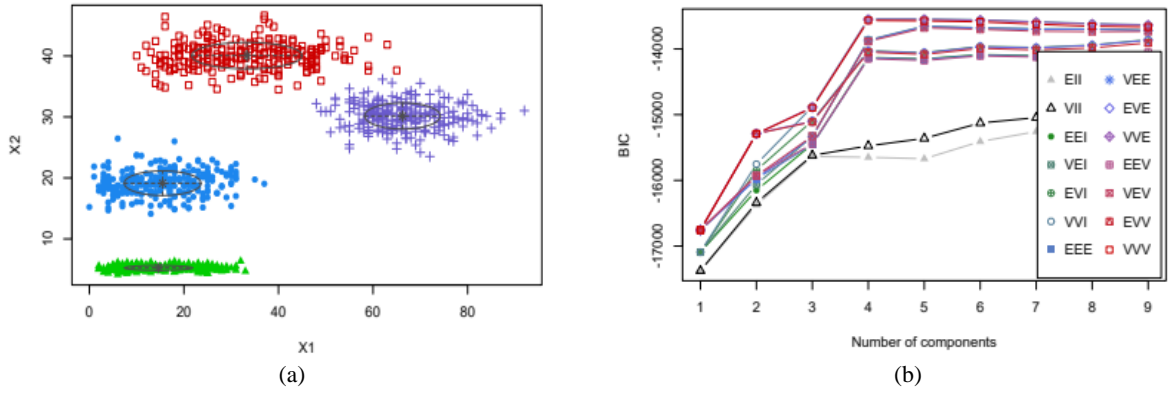

(a)          (b)

**FIGURE 10.** BIC diagram and the clustering results of GMM using 4 factors.

Finally, figure 11 shows the outcome of the hierarchical agglomerative clustering algorithm UPGMA. The tree diagram is cut at a suitable point aiming to produce 4 clusters.

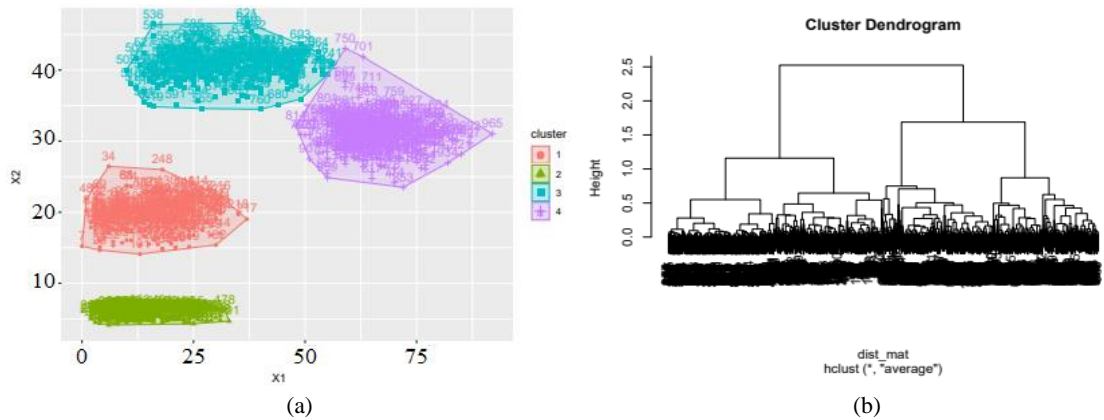

(a)          (b)

**FIGURE 11.** UPGMA clustering results accompanied by the respective dendrogram.

## Clustering Performance Examined on Continuous Observations

The second simulated dataset is characterized by 2-dimensional r.v. following normal (Gaussian) distributions. We produce 1000 continuous observations constituting a total of 5 classes (Figure 12).

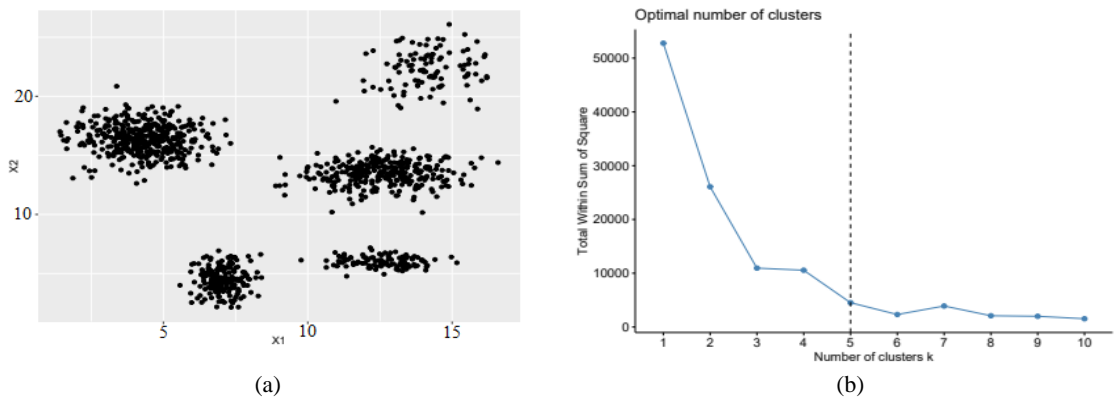

(a)          (b)

**FIGURE 12.** Diagrammatic representation of dataset 2 and the corresponding WCSS graph that aid the recognition of the appropriate number of clusters.

Figure 12 shows a rapid decrease in the WCSS curve as $k$ increases. For $k \geq 5$, we observe the emergence of a plateau. Therefore, the elbow method suggests the usage of five initial medoids. Using the OPTICS algorithm, we observe a total of 5 prevalent peaks, resulting in an equal number of valleys, indicating the existence of 5 classes (Figure 13).

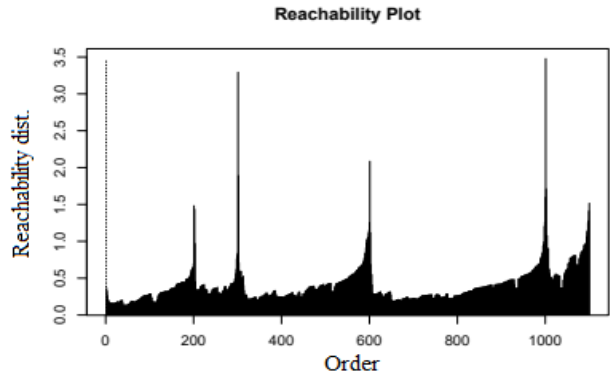

**FIGURE 13.** Reachability plot for dataset 2.

Initially, the naive K-means is applied (Figure 14), where the two classes on the right upper part are merged into one. In addition, from the one class located in the lower left part of the graph, two clusters have been formed. The inability of the K-means to create appropriate clusters has been also presented in the previous application. In contrast, K-means++ achieves a better classification efficiency. Although, like the conventional K-means, it splits a unique class into two clusters while forming an incorrect cluster (red color) containing observations of two other groups.

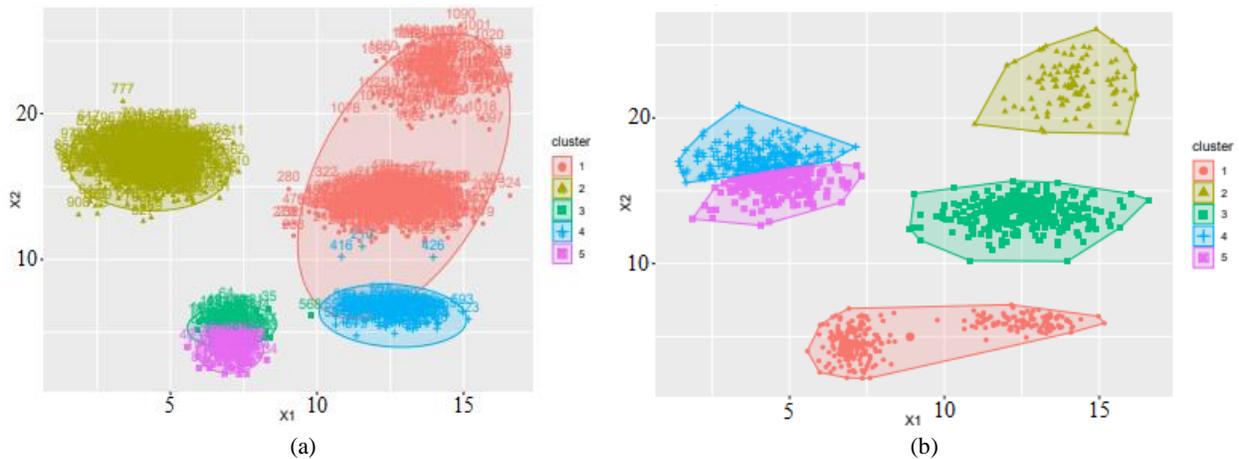

(a)          (b)

**FIGURE 14.** Clustering of dataset 2 based on K-means and K-means++.

An embodiment of the elbow method for the case of PAM is shown in figure 15. The selection of 5 representative centers seems to remain the most suitable option, a phenomenon that is quite expected. Figure 15 also shows the clustering efficacy of PAM, where it appears to be a complete match between classes and clusters.

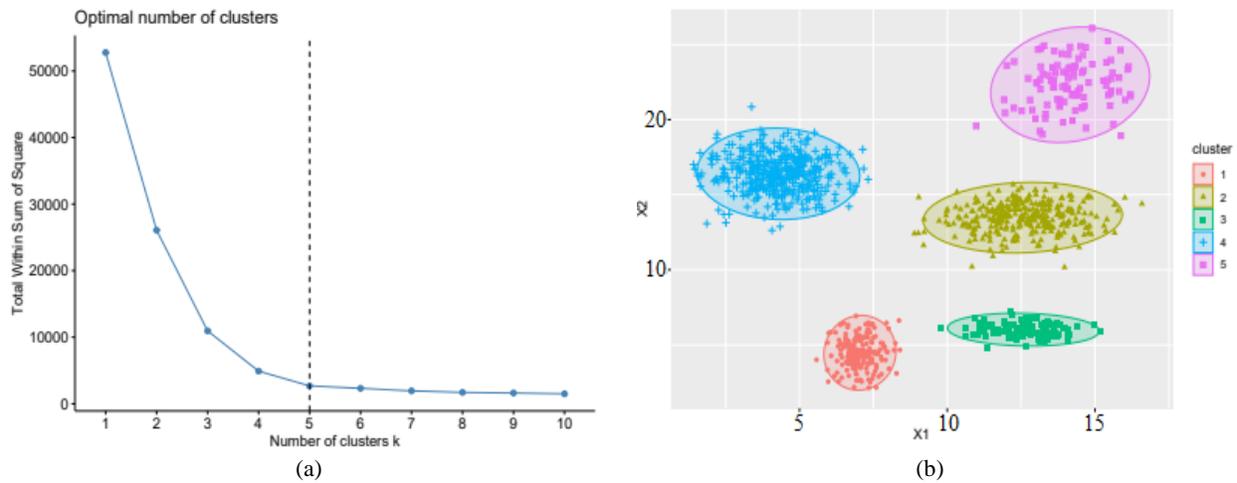

**FIGURE 15.** Elbow Method and clustering outcome for the PAM algorithm.

Furthermore, figure 16 represents the application of CLARA on 50 samples of a total of 50 observations. We observe that the large number of samples yields a rarely satisfactory result, with a relatively reduced computational time, as we discuss later.

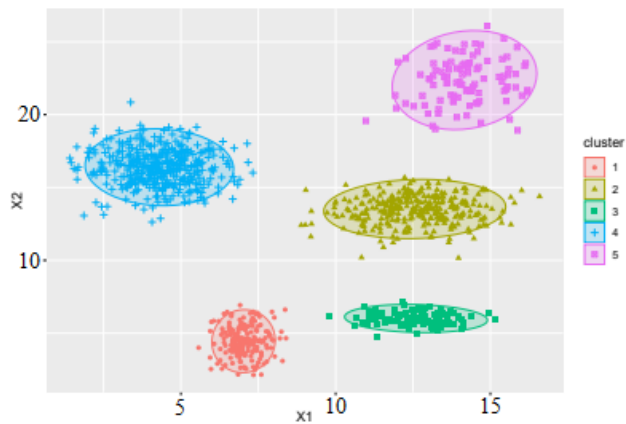

**FIGURE 16.** CLARA's clustering results.

Figure 17 displays the elements' distances from their 3rd, 4th and 5th closest neighbor respectively. The parameter $\varepsilon$ is estimated around 0.6 for all three cases. Additionally figure 17 displays the formation of additional clusters with negligible number of components and a significantly large number of irregular points during the implementation of DBSCAN. For $minPts = 5$, the previous fact seems to be changing. The extra clusters have disappeared and the correspondence between classes and clusters seems sufficient.

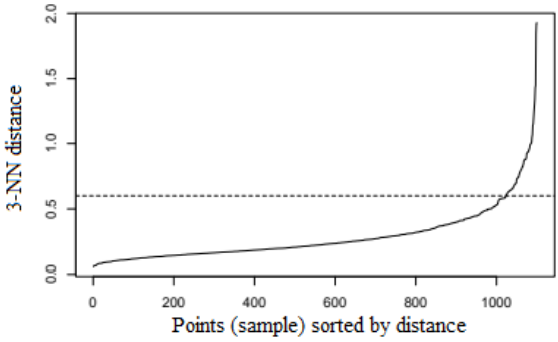
(a)

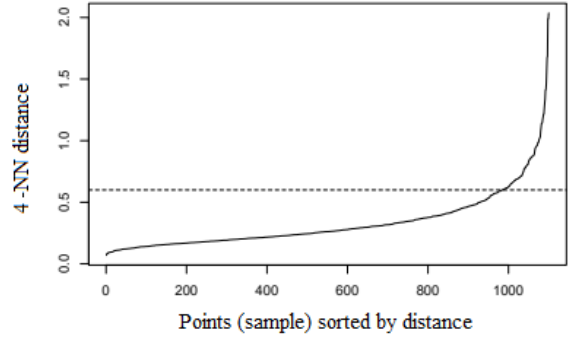
(b)

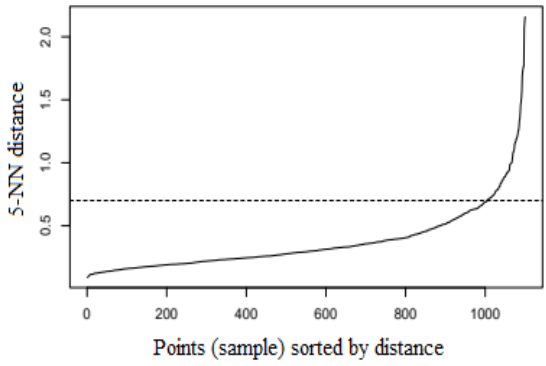
(c)

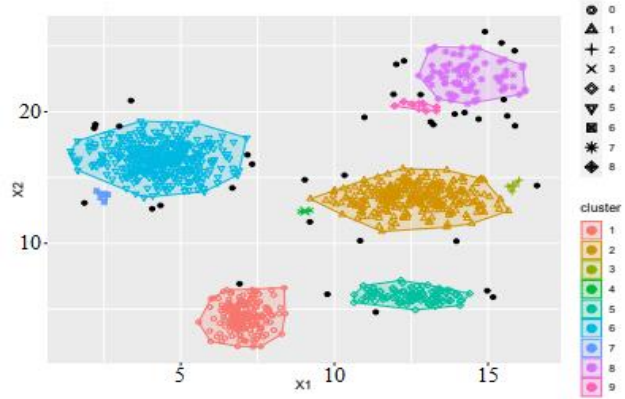
(d)

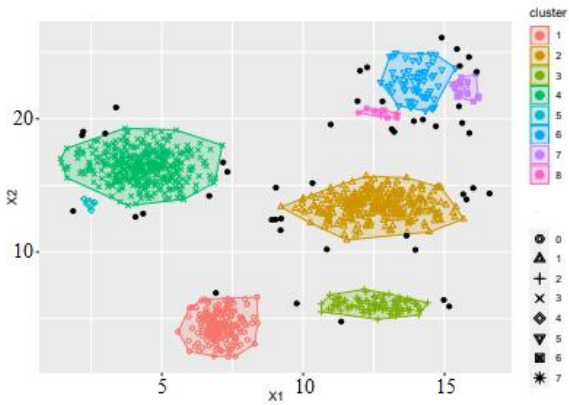
(e)

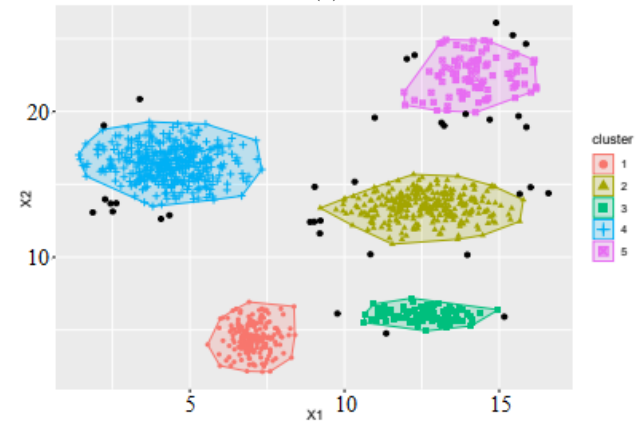
(f)

**FIGURE 17.** Graph for dataset points' distances from their 3rd, 4th and 5th nearest neighbour and the DBSCAN grouping results for (minPts = 3, $\varepsilon$ = 0.6), (minPts = 4, $\varepsilon$ = 0.6) and (minPts = 5, $\varepsilon$ = 0.6).

As previously, we explore the formed clusters that arise utilizing the OPTICS's reachability plot. We choose a value for the reachability distance approximately equal to the lowest of the five prevalent peaks. This value set the boundaries between the cluster formation and the noise points that may occur (Figure 18).

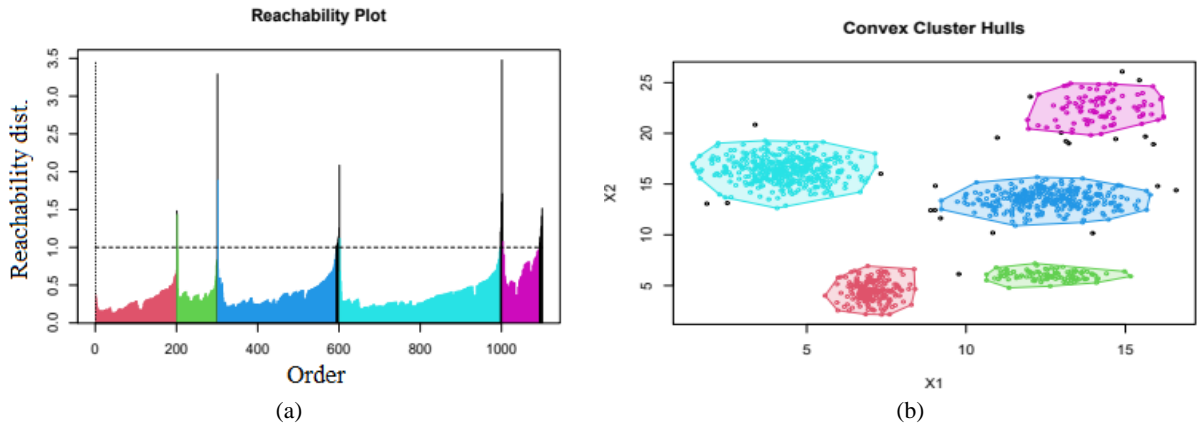

**FIGURE 18.** Reachability plot accompanied by the clustering of OPTICS for dataset 2 using distance = 1.

Figure 19 shows the application of BIC (26) on the observations of the 2nd dataset. The displayed legend concerns the shape of the distributions, such as the area occupied by the ellipsoid and its orientation, information resulting from the assessment of the covariance matrix. There is a clear indication that the GMM should be composed of 5 factors for which the extent and shape of the ellipsoid differ but the orientation in all cases is set in accordance with the axes. Also, figure 19 displays the GMM's results for the VVI case, which stands for the diagonal and varying volume and shape of each ellipse, while figure 20 is dedicated to UPGMA's grouping outcome.

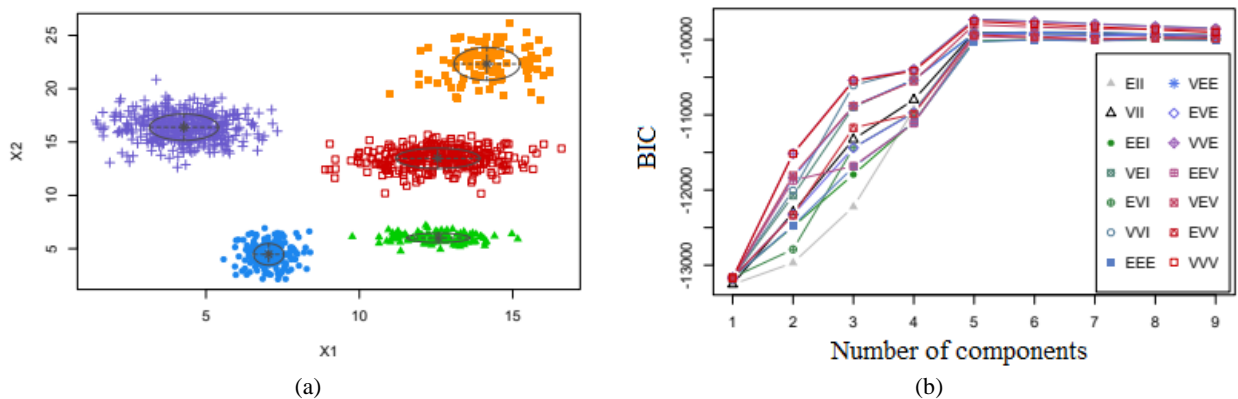

**FIGURE 19.** BIC diagram and clustering of the 5 classes according to GMM for dataset 2.

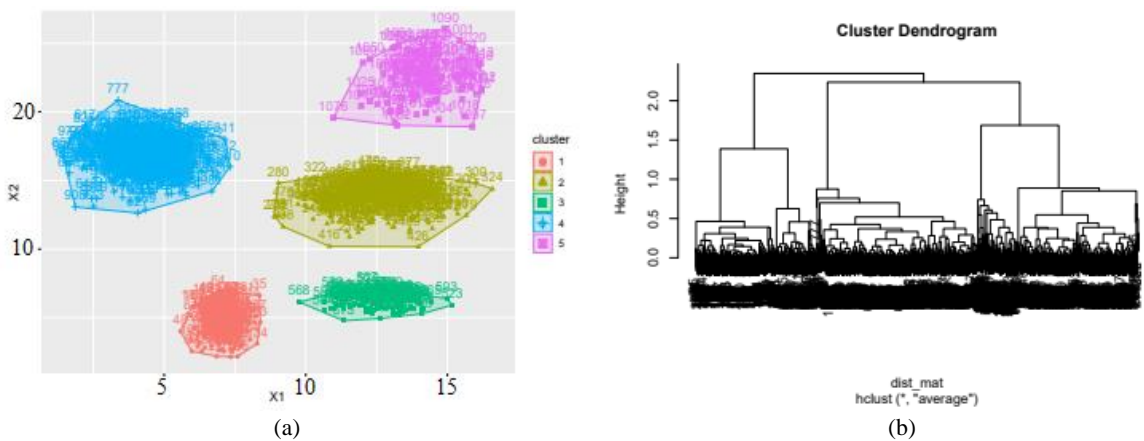

**FIGURE 20.** UPGMA clustering results and the respective dendrogram applied on dataset 2.

## Clustering Performance Examined on Discrete Observations

The observations of the third dataset are derived from the utilization of two Poisson distributions, culminating in 1000 discrete observation values (Figure 21). The data points are generated with the prospect of creating 4 separate classes. However, the produced classes are quite close, complicating the clustering procedure significantly. The spatial density of observations does not seem to vary significantly from region to region; hence it is not expected to encounter serious obstacles in the parameter estimation of DBSCAN. Using the elbow method (Figure 21), we notice a rapid drop in WCSS until $k = 4$ medoids, signifying the emergence of a plateau.

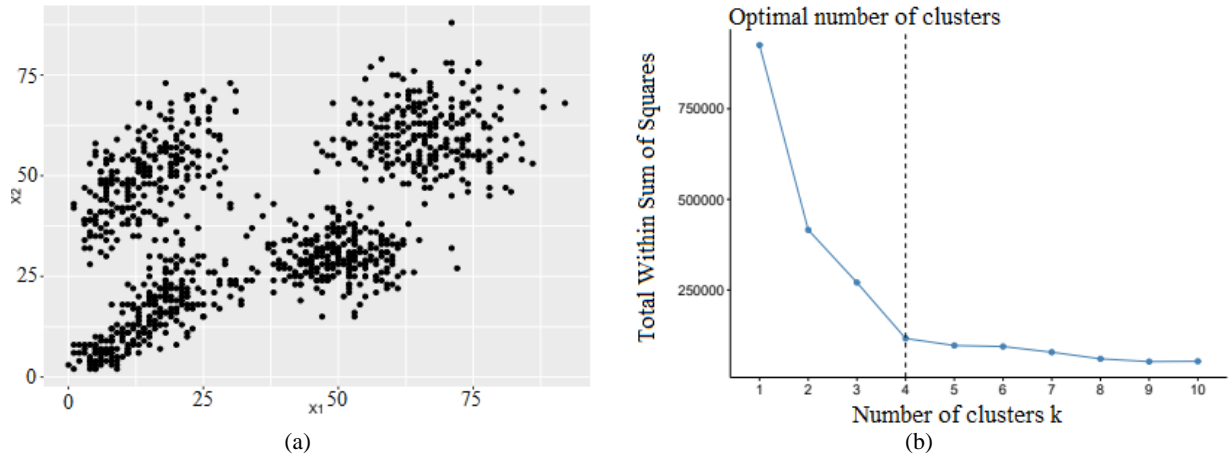

**FIGURE 21.** Observations of dataset 3 with the respective WCSS diagram.

Using 4 clusters as an input for K-means and K-means++ (Figure 22), we accomplish a relatively robust identification of the respective classes. The percentage of the correct classifications is shown in Table 3 (conclusions) in parallel with their required convergence times.

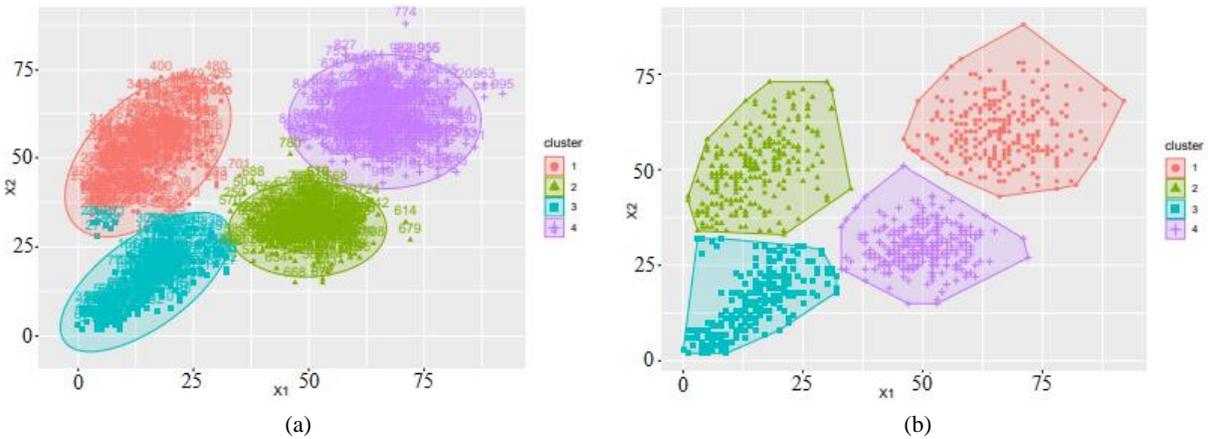

**FIGURE 22.** K-means and K-means++ clustering for dataset 3.

The result of K-means++ appears to be slightly better than that of K-means, while the necessary comparisons are made later when the appropriate tables are given. Figure 23 represents the results of the application of PAM and CLARA respectively on dataset 3, that seem to be highly similar, while CLARA displays slightly increased number of outliers.

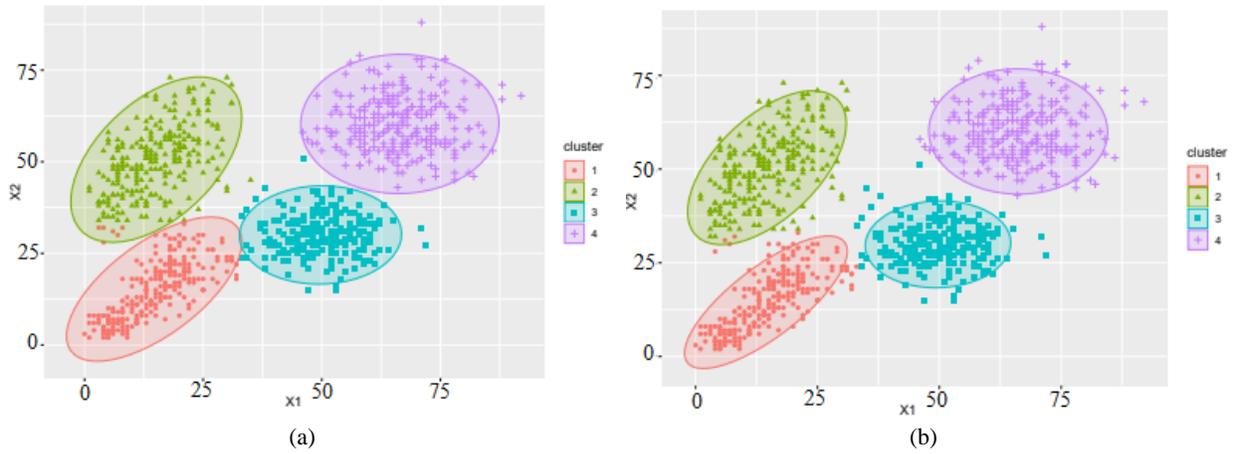

**FIGURE 23.** PAM and CLARA results for dataset 3.

Figure 24 presents the plots of the dataset points' distances from their 3rd, 4th, and 5th neighbor. The application of DBSCAN for $(minPts = 3, \varepsilon = 2.5), (minPts = 4, \varepsilon = 3)$ and $(minPts = 5, \varepsilon = 3)$ is displayed in the same figure, where the main body of each class has been successfully recognized in all occasions. In the first 2 cases, except from the 4 main clusters and the irregular points, extra clusters with negligible populations are created. These additional clusters do not significantly affect the percentage of correct classifications, while a different parameter selection might confront this problem. The fact that additional classes are created, in addition to those expected from visual examination, is anticipated since DBSCAN does not require an initialization for the number of clusters.

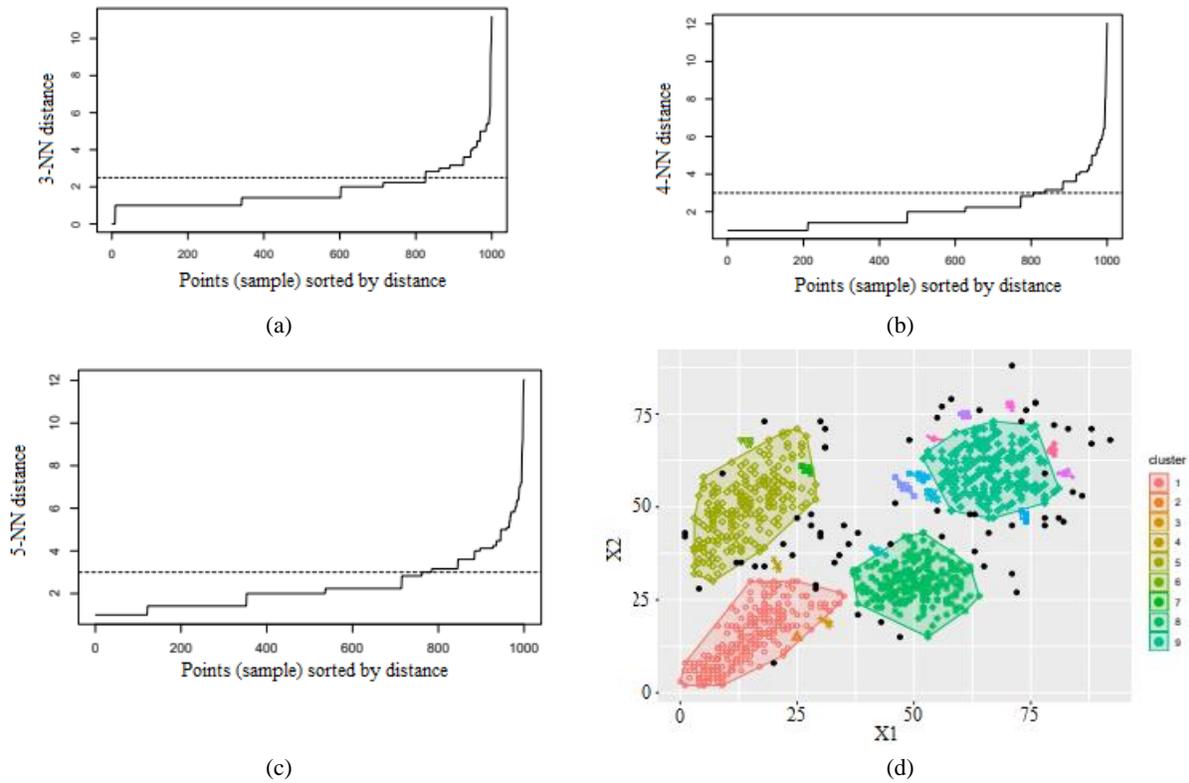

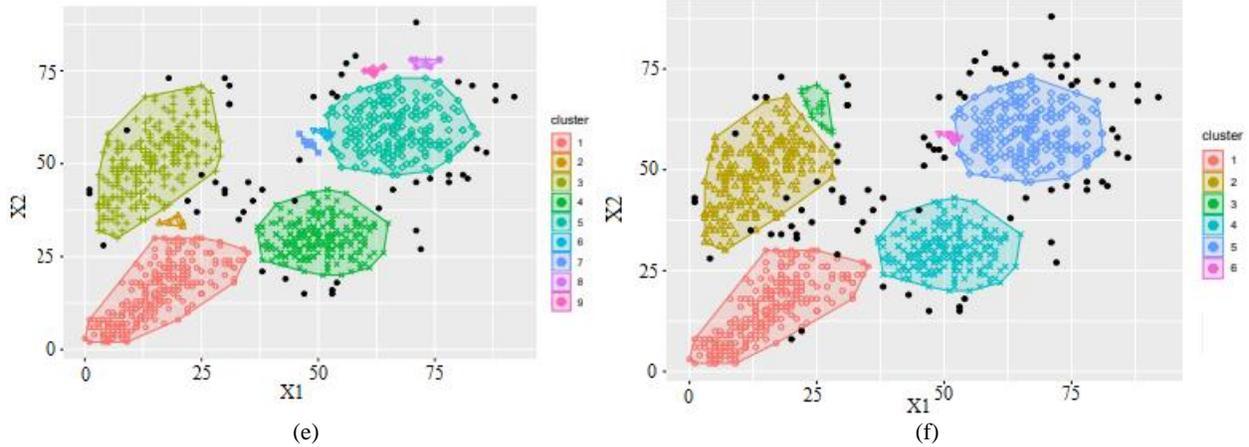

(e)                                         (f)

**FIGURE 24.** Graph for the 3rd, 4th and 5th nearest neighbour and DBSCAN application using (minPts = 3, $\varepsilon$ = 2.5), (minPts = 4, $\varepsilon$ = 3) and (minPts = 4, $\varepsilon$ = 3) on dataset 3.

The reachability plot shown in figure 25, displays 4 high peaks that stand out underlining the confirmation of the existence of 4 clusters. For the reachability distance we set a value of 6.5 and the suggested clustering selection is introduced with 4 respective colors. The black points will not be assigned to any cluster. Notice that through the usage of OPTICS, we avoid the creation of clusters with negligible sizes in contrast to the grouping proposed by DBSCAN.

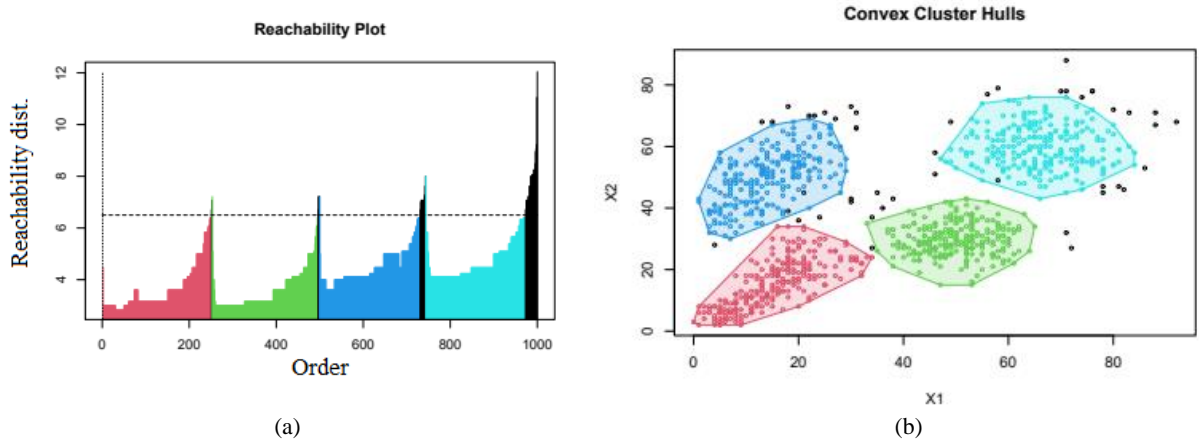

(a)                                         (b)

**FIGURE 25.** Reachability plot for dataset 3 accompanied by OPTICS's clustering results.

There is an indication that the GMM should consist of 5 factors. However, the BIC's value for 4 factors is quite close to the maximum value, while considering the selection of $k$ for the previous results, we culminate in setting $k = 4$. In this case, the ellipsoid's orientation differs. The 2nd part of figure 26, presents the results of the GMM, while figure 27 provides the produced grouping based on UPGMA method.

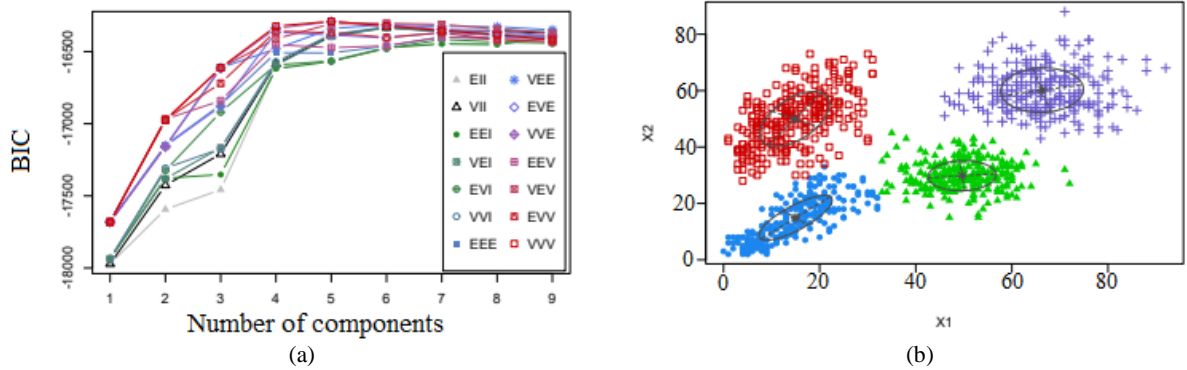

**FIGURE 26.** BIC values for various initializations of the GMM's parameters and the respective clustering results.

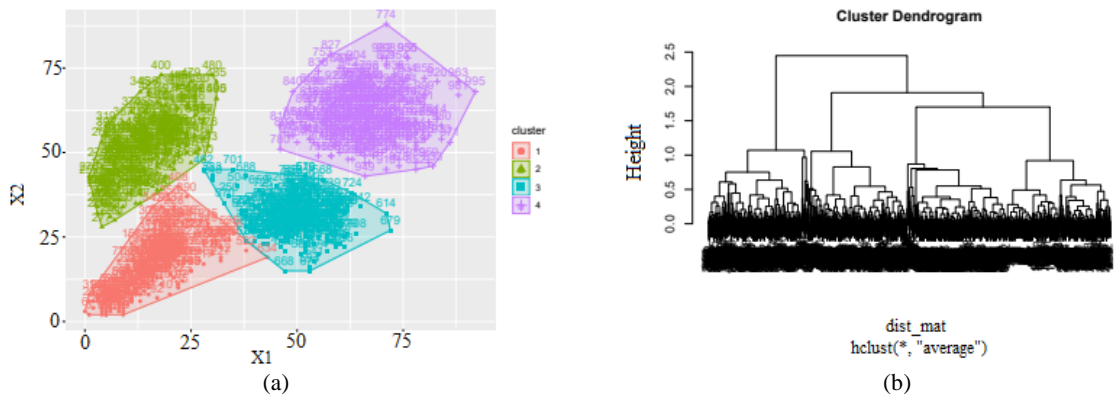

**FIGURE 27**. UPGMA clustering results and the respective dendrogram.

## DISCUSSION

In this part we discuss the produced results, concerning the clustering efficacy and running time of the examined methodologies on the simulated datasets. Regarding the 3 explored simulated datasets, figures 28, 29 and 30 present the required convergence times (Running Times) accompanied by the respective accuracies for each of the utilized clustering algorithms. In all cases the same computer system is used.

It can be observed that the corresponding times for the 2nd dataset are usually longer than those of the other 2 sets, highlighting that the continuous observations result into more computationally expensive operations. K-means seems to be the fastest in all cases. This is not a paradox since the fast convergence of this algorithm represents one of its characteristic features. The accuracy of K-means for the 1st dataset is by far the worst compared to all other methodologies. As mentioned in the 3.2 paragraph, this is a significant disadvantage of the naive K-means that derives from the random selection of the initial medoids.

| Clustering Algorithm/feature | Running Time (msecs) | Accuracy |
|---|---|---|
| GMM | 596.8147 | 0.998 |
| k-means | 2.05315 | 0.699 |
| kmeans++ | 237.6812 | 0.976 |
| Optics | 12.25786 | 0.992 |
| Clara | 25.98713 | 0.971 |
| Pam | 292.6252 | 0.976 |
| UPGMA | 63.6253 | 0.995 |
| Dbscan, MinPts=3 | 4.606563 | 0.944 |
| Dbscan, MinPts=4 | 3.308793 | 0.944 |
| Dbscan, MinPts=5 | 2.93321 | 0.949 |

**FIGURE 28**. Dataset 1.

| Clustering Algorithm/feature | Running Time (msecs) | Accuracy |
|---|---|---|
| GMM | 764.8325 | 1.000 |
| k-means | 1.8484 | 0.7391 |
| kmeans++ | 353.8694 | 0.7389 |
| Optics | 59.96409 | 0.9891 |
| Clara | 30.94512 | 1.000 |
| Pam | 366.0111 | 1.000 |
| UPGMA | 73.0520 | 1.000 |
| Dbscan, MinPts=3 | 6.128623 | 0.9355 |
| Dbscan, MinPts=4 | 3.42663 | 0.9509 |
| Dbscan, MinPts=5 | 3.071583 | 0.9689 |

**FIGURE 29**. Dataset 2.

| Clustering Algorithm/feature | Running Time (msecs) | Accuracy |
|---|---|---|
| GMM | 658.9313 | 0.991 |
| k-means | 2.1453 | 0.982 |
| kmeans++ | 220.2396 | 0.982 |
| Optics | 14.97722 | 0.953 |
| Clara | 24.53058 | 0.988 |
| Pam | 242.6905 | 0.983 |
| UPGMA | 43.54115 | 0.984 |
| Dbscan, MinPts=3 | 3.75497 | 0.938 |
| Dbscan, MinPts=4 | 3.007463 | 0.929 |
| Dbscan, MinPts=5 | 5.66411 | 0.926 |

**FIGURE 30**. Dataset 3.

K-means++ for different datasets is clearly slower than the naive K-means. Although the accuracy for the 1st dataset is significantly better. Regarding the other 2 sets, the accuracy remains at the levels of the simple algorithm. PAM in all cases achieves highly satisfactory accuracy especially in the case of the 2nd set with perfect clustering efficiency. The respective running time is the 2nd longest compared to all 3 sets. While this is not a problem for the specific amount of data, its application to larger sets may force someone to turn to CLARA's fastest alternative.

Regarding CLARA, we observe clearly shorter convergence times, which is quite expected due to the algorithm's operation. We previously mentioned that during the application of CLARA, the PAM methodology is used to create clusters on samples of the original dataset, making CLARA significantly faster. Now the important issue is its final accuracy. The application of the algorithm to samples of appropriate size, seems to produce reliable grouping results since performance equivalent to that of PAM is achieved. An alternative to even faster convergence is the usage of smaller samples at the risk of deteriorating the accuracy.

The convergence time of OPTICS in the case of the 2nd dataset, appears to be quite longer than the 1st one. This fact derives from the larger fluctuations of the densities inside the 2nd dataset's clusters. The accomplished accuracy after a specific selection of the reachability distance is in all cases satisfactory. In the case of DBSCAN, the running times are generally at the same levels for all sets. The difference in time with OPTICS, derives from the fact that OPTICS is considered to manage a wider range of parameter values rather than a universal value for the whole dataset. To clarify that, the running time of OPTICS includes the calculation of the reachability plot. The irregular points produced from these two algorithms are considered as misclassifications.

The required time for the convergence of the expectation maximization algorithm – that is responsible for the estimation of the suitable GMM – is the highest for all 3 sets. In fact, in dataset 2, the necessity of additional parameter estimation culminates in a higher computational cost. For all 3 datasets, the accuracy is one of the best among the tested algorithms. Finally, the UPGMA method displays quite trustworthy efficiency, while the respective running times are a bit slower.

The clustering efficiency of the conventional K-means seems to be significantly surpassed by more robust methodologies, while its utilization should be restricted in cases where we are satisfied with a hasty grouping procedure. The GMM method displayed nearly perfect accuracy in all 3 cases; although his increased running time may constitute an inhibitory factor for larger datasets. Hence, the usage of GMMs may be considered as more suitable in cases of smaller datasets or datasets of highly complicated cluster shapes. The best balance between clustering accuracy and convergence time is provided by CLARA algorithm, presenting valuable indications about its appropriateness in computationally expensive sets. Finally, in cases where the identification of irregularities is necessary, DBSCAN's particularities should be proven as beneficial, while OPTICS would be able to cope with more challenging sets.

## CONCLUSIONS

This paper provides a detailed presentation of the most widely used clustering methods, giving information not only for the operation and the methodology behind these algorithms but also for their clustering efficiency and running time. For each algorithm we provide valuable information about the ideal parameter selection to enhance their grouping ability and efficacy, while the reference in each algorithm's complexity displays important elements about their appropriateness in accordance with the dataset's size.

The exploration of the accuracy and the running time on three datasets containing discrete, continuous observations and a mixture of them, reveals useful conclusions about the suitability of each algorithm depending on the observations' nature. As expected, the trustworthiness of the naive K-means, seems to be highly surpassed by subsequent clustering methodologies. The GMM algorithm displays the best grouping efficiency in all cases, although having a much longer converge time due to its way of operation, deeming this methodology ideal for more complex datasets. The most suitable balance between accuracy and convergence period for larger datasets may be produced by the CLARA and PAM algorithms, while quite suitable balance is also provided by UPGMA hierarchical clustering methodology with a bit longer running time. In cases where the identification of outliers is necessary, DBSCAN and especially OPTICS constitute an ideal selection.